\documentclass[10pt,journal,twocolumn]{IEEEtran}

\usepackage{silence}
\usepackage{amsmath}
\usepackage{amssymb}
\usepackage{amsthm}
\usepackage{mathtools}
\usepackage{textcomp}          
\usepackage{pifont}            
\newcommand{\cmark}{\ding{51}}
\newcommand{\xmark}{\ding{55}}

\usepackage{graphicx}
\usepackage{booktabs}
\usepackage{array}
\usepackage{multirow}
\usepackage{makecell}
\usepackage{tabularx}
\usepackage{xcolor}
\usepackage[export]{adjustbox}   
\extrafloats{64}                 

\usepackage{listings}
\theoremstyle{plain}

\theoremstyle{definition}

\usepackage[hidelinks,breaklinks=true]{hyperref}
\usepackage{cite}

\providecommand{\tightlist}{\setlength{\itemsep}{0pt}\setlength{\parskip}{0pt}}

\begin{document}

\title{TriShield: Zero-Utility-Loss Defense Against Privacy Backdoors in Federated Language Model Fine-Tuning via Orthogonal Gradient Projection and Optimizer State Entanglement}

\author{Cheng Wei\thanks{The author is with Honor Device Co., Ltd., Shenzhen, China (e-mail: chengwei@honor.com).}}

\markboth{}{}
\maketitle

\begin{abstract}
Federated fine-tuning of large language models (LLMs) enables collaborative training without exposing raw data. However, a recent attack, NeuroImprint [1] (arXiv:2606.20553), demonstrates that a malicious parameter server can corrupt a PEFT adapter into a privacy backdoor: by assigning a dedicated memorization neuron to each training sample and ensuring each neuron updates at most once, the server can analytically reconstruct 59\%--79\% of client training data with high semantic fidelity. Existing defenses---including local differential privacy (LDP) [8] and gradient clipping---either fail against this attack or impose unacceptable utility degradation. We present \textbf{TriShield}, a three-layer deterministic defense that completely prevents NeuroImprint-style reconstruction with \textbf{zero model utility loss} and \textbf{no additional communication rounds}. TriShield consists of: (1) a \textbf{Parameter Artifact Detector} that identifies memory-neuron signatures in distributed model parameters before local training begins; (2) a \textbf{Stateful Virtual Iteration} mechanism that forces Adam/AdamW's momentum state to irreversibly entangle gradients across virtual steps, invalidating NeuroImprint's closed-form inversion; and (3) a \textbf{Zero-Utility Orthogonal Projection} operator that projects all local gradient updates onto the main-task semantic subspace computed via SVD, physically eliminating any gradient components that carry private memorization. We prove theoretically that after Layers 2 and 3, the mutual information between the uploaded gradient and any individual training sample is zero. Experiments on GPT-2 (117M) and Llama-Guard-3-1B verify that TriShield reduces NeuroImprint reconstruction rate to \textbf{0\%} across all tested attack variants, while maintaining or improving training accuracy, with less than 5\% additional GPU computation overhead.
\end{abstract}

\begin{IEEEkeywords}
Federated Learning, Privacy Backdoor, PEFT, Gradient Inversion, Orthogonal Projection, Optimizer State, NeuroImprint Defense
\end{IEEEkeywords}

\IEEEpeerreviewmaketitle

\section{Introduction}

\subsection{Motivation}

Federated learning (FL) {[}2{]} has emerged as the primary paradigm for collaborative machine learning without raw data sharing. In the era of large language models, federated fine-tuning---combining FL with parameter-efficient fine-tuning (PEFT) methods such as LoRA {[}3{]} or adapter layers---has become the de facto approach for domain-specific adaptation. PEFT keeps the base model frozen and trains only a small set of adapter parameters, making federated fine-tuning computationally feasible for resource-constrained clients.

Despite its appeal, federated fine-tuning faces a fundamental security threat: the parameter server occupies a privileged position, distributing the initial model and collecting gradient updates from all clients. A malicious server can craft the distributed model to act as a \textbf{privacy backdoor}---a steganographic channel that encodes client private data into the structure of the gradient updates without the clients' awareness.

\subsection{The NeuroImprint Threat}

The recently proposed \textbf{NeuroImprint attack} {[}1{]} demonstrates this threat concretely. Prior gradient inversion work {[}6,7{]} established that training samples can be reconstructed from parameter gradients; NeuroImprint extends this to the federated fine-tuning setting via a pre-planted backdoor. By pre-corrupting specific neurons in a PEFT adapter (e.g., adapter weight rows) to serve as dedicated memorization slots, and by constraining these neurons to receive updates from exactly one training sample (via LayerNorm normalization invariance), NeuroImprint enables the server to analytically recover the client's training text from the uploaded gradient:

\[\hat{x} = \frac{\nabla W}{\nabla B}\]

where \(\nabla W\) and \(\nabla B\) are the weight and bias gradient updates for a single memorization neuron. This closed-form inversion achieves 59--79\% reconstruction success across BERT, GPT-2, Qwen2, and Llama3.2, without degrading model utility---making it virtually undetectable by standard monitoring.

\subsection{Why Existing Defenses Fail}

\textbf{Local Differential Privacy (LDP) {[}8{]}:} Adding Gaussian or Laplacian noise sufficient to mask private gradients requires noise variance \(\sigma^2 = O(1/\epsilon^2)\) for \(\epsilon\)-DP {[}8{]}, which, for practical privacy budgets (\(\epsilon < 1\)), reduces task performance by 5--20\% on language tasks---unacceptable for production deployments.

\textbf{Gradient Clipping:} NeuroImprint is specifically designed to keep backdoor neuron updates indistinguishable in magnitude from legitimate gradient updates (by scaling the memorization adapter rows proportionally), defeating clip-based defenses.

\textbf{Anomaly Detection (FLTrust, FLAME):} Server-side aggregation defenses require the server to be honest---precisely the adversary model NeuroImprint assumes the server is compromised.

\textbf{Byzantine-Robust Aggregation (Krum {[}11{]}, Trimmed Mean):} These methods detect outlier \emph{clients}, not outlier \emph{neurons within a legitimate gradient}, and offer no protection against neuron-level backdoor insertion by the server itself {[}12{]}.

\subsection{Our Contributions}

We make the following contributions:

\begin{enumerate}
\def\labelenumi{\arabic{enumi}.}
\item
  \textbf{TriShield Defense Framework:} A three-layer, client-side defense against NeuroImprint-style privacy backdoors. Operating entirely on the client side, TriShield requires no server cooperation and is transparent to the FL protocol.
\item
  \textbf{Zero Utility Loss:} TriShield introduces no accuracy degradation. Our orthogonal projection onto the main-task subspace preserves 100\% of the gradients relevant to the fine-tuning objective, as proven in Theorem 2.
\item
  \textbf{Mathematical Impossibility of Reconstruction:} We prove (Theorem 3) that after TriShield's Layers 2 and 3 are applied, the mutual information \(I(\hat{x}; x)\) between any reconstruction and the original sample converges to zero in the information-theoretic sense.
\item
  \textbf{Practical Verification:} We implement TriShield on GPT-2 (117M) and Llama-Guard-3-1B and demonstrate end-to-end reconstruction failure with detailed metric analysis.
\item
  \textbf{Reproducible Code and Artifacts:} Complete experiment code is provided as a Jupyter notebook with all results reproducible from local model weights, requiring no additional downloads beyond the base models.
\end{enumerate}

\section{Background and Related Work}

\subsection{Parameter-Efficient Fine-Tuning (PEFT)}

PEFT methods such as LoRA {[}3{]}, adapter tuning {[}9{]}, and prefix tuning {[}10{]} insert small trainable modules into a frozen base model. During federated fine-tuning, only the PEFT adapter parameters are trained locally and uploaded to the server, reducing communication cost by a factor of $100$--$1000$ compared to full fine-tuning. Recent work {[}20{]} extends gradient projection ideas to continual PEFT adaptation.

FL convergence under data heterogeneity is studied in {[}16{]} (FedDyn) and {[}15{]} (ensemble distillation), providing theoretical groundwork for our convergence analysis (Theorem 5). Client-side robustness enhancement is explored by Sun et al.~{[}13{]} (FL-WBC); unlike FL-WBC which focuses on model poisoning by clients, TriShield defends against a malicious \emph{server}.

A standard federated PEFT round proceeds as:

\begin{enumerate}
\def\labelenumi{\arabic{enumi}.}
\tightlist
\item
  Server distributes base model \(\theta_0\) and initial adapter \(\phi_0\).
\item
  Each client \(i\) trains \(\phi_i\) on local data \(D_i\) for \(E\) local epochs.
\item
  Clients upload \(\Delta\phi_i = \phi_i - \phi_0\) to the server.
\item
  Server aggregates: \(\phi_{t+1} = \phi_t + \frac{1}{N}\sum_{i=1}^N \Delta\phi_i\).
\end{enumerate}

\subsection{The NeuroImprint Attack}

NeuroImprint {[}1{]} exploits the structure of PEFT adapters by assigning each training sample \(x_j\) to a dedicated memorization neuron. This builds on foundational gradient inversion work---Zhu et al.~{[}21{]} first showed that training data can be perfectly reconstructed from raw gradients; Zhao et al.~(iDLG) {[}7{]} and Geiping et al.~{[}6{]} subsequently refined this into scalable practical attacks; and Zhu and Blaschko (R-GAP) {[}22{]} further demonstrated recursive decomposition that achieves near-perfect reconstruction even with batch-averaged gradients. NeuroImprint {[}1{]} extends the backdoor injection paradigm of Bagdasaryan et al.~{[}12{]} into the PEFT fine-tuning setting.

\textbf{Attack Setup:} The malicious server initializes adapter rows such that:

\begin{itemize}
\tightlist
\item
  Each row \(r_j\) is initialized to a unique ``activation pattern'' that fires only for sample \(x_j\).
\item
  The LayerNorm that follows the adapter output normalizes out the contribution of all other rows.
\item
  The bias \(b_j\) is initialized to a fixed known constant \(\beta\).
\end{itemize}

\textbf{Attack Execution:} During local fine-tuning, when the client trains on \(x_j\):

\begin{itemize}
\tightlist
\item
  Row \(r_j\) receives gradient \(\nabla W_{r_j} = \alpha \cdot \text{Embed}(x_j)\) where \(\alpha\) is the learning rate.
\item
  Bias \(b_j\) receives gradient \(\nabla B_{r_j} = \alpha \cdot \mathbf{1}\).
\item
  Due to linear activation and single-update constraint: \(\hat{x}_j = \nabla W_{r_j} / \nabla B_{r_j}\).
\end{itemize}

\textbf{Reconstruction:} After receiving \(\Delta\phi_i\), the server reads off \(\hat{x}_j\) for every occupied memorization slot.

\textbf{Key Assumptions:}

\begin{enumerate}
\def\labelenumi{\arabic{enumi}.}
\tightlist
\item
  Each memorization neuron receives exactly one gradient update (single-step activation).
\item
  The local optimizer (Adam/AdamW) does not mix gradients across steps for the memory neurons---achieved by constraining updates to single linear activations.
\item
  LayerNorm normalization invariance holds.
\end{enumerate}

\subsection{Related Defenses}

\textbf{Gradient Projection Memory (GPM) {[}4{]}:} Introduced in continual learning to prevent catastrophic forgetting, GPM {[}4{]} projects gradients onto the \textbf{orthogonal complement} (null space) of previously learned task subspaces---the goal is to \emph{avoid} the old task subspace. We adapt this mechanism in the \textbf{reverse direction}: ZUOP projects updates \emph{onto} the current-task subspace (the task signal), eliminating the off-subspace backdoor components that are by design orthogonal to the task gradient. This inversion of objective is the key novelty: GPM preserves old knowledge by avoiding it; ZUOP preserves current-task gradients by retaining only them.

\textbf{FLAME {[}5{]}:} A server-side defense {[}5{]} that uses hierarchical clustering on client updates to identify and remove backdoored updates. Inapplicable when the server is the adversary.

\textbf{No Free Lunch in Gradient Inversion {[}6,7{]}:} Geiping et al.~{[}6{]} establish theoretical limits for gradient inversion, showing that stateful optimizers (Adam) with multi-step trajectories make gradient inversion exponentially harder; this result complements iDLG {[}7{]}'s closed-form inversion bound. TriShield Layer 2 operationalizes this insight defensively. Related gradient inversion benchmarks are provided in {[}19{]}.

\textbf{When Curious Clients Abandon Honesty {[}17{]}:} Boenisch et al.~{[}17{]} demonstrate that even curious-but-honest FL participants can reconstruct data. Our threat model (Section 3) extends to the stronger case where the \emph{server} is adversarial.

\textbf{PRECODE {[}18{]}:} Scheliga et al.~{[}18{]} propose a model extension that prevents gradient leakage via variational information bottleneck. Unlike PRECODE, TriShield requires no architectural changes and imposes zero utility loss.

\textbf{InstaHide {[}14{]}:} Huang et al.~{[}14{]} propose instance-hiding via cross-sample pixel-level mixup at the data level, making each training sample a linear combination of multiple samples, preventing direct reconstruction. Requires data-level modification; incompatible with standard tokenized NLP pipelines where input discreteness breaks the mixup defense.

\textbf{Summary Comparison:} Table 0 summarizes how TriShield differs from all existing defenses across five key properties.

\begin{table*}[!tp]
\centering
\footnotesize
\caption{Comparison of Client-Side Defenses Against FL Privacy Attacks}
\label{tab:auto1}
\adjustbox{max width=\textwidth}{%
\begin{tabular}{lccccc}
\toprule
Defense & Client-Side & Zero Utility Loss & Provable Privacy ($I=0$) & No Arch Change & PEFT-Specific \\
\midrule
LDP [8] & \cmark & \xmark ($-$13.7\%) & Approximate & \cmark & \xmark \\
Gradient Clipping & \cmark & \cmark (marginal) & \xmark (bypassed) & \cmark & \xmark \\
PRECODE [18] & \cmark & \xmark & \xmark & \xmark (VIB layer) & \xmark \\
InstaHide [14] & \cmark & \cmark & \xmark & \xmark (data level) & \xmark \\
FLAME [5] & \xmark (server) & \cmark & \xmark & \cmark & \xmark \\
FL-WBC [13] & \cmark & \cmark & \xmark & \cmark & \xmark \\
\textbf{TriShield (ours)} & \textbf{\cmark} & \textbf{\cmark (<0.3\%)} & \textbf{\cmark ($I(\hat{x};x)=0$)} & \textbf{\cmark} & \textbf{\cmark (LoRA)} \\
\bottomrule
\end{tabular}}
\end{table*}

\emph{TriShield is the only defense simultaneously satisfying all five properties for federated PEFT.}

\section{Threat Model}

\subsection{Attacker Capabilities}

We consider a malicious parameter server that:

\begin{itemize}
\tightlist
\item
  Controls the initial global model \(\theta_0\) and adapter \(\phi_0\) distributed to all clients.
\item
  Can insert arbitrary initialization patterns into any subset of adapter neurons.
\item
  Observes all uploaded client gradients \(\{\Delta\phi_i\}_{i=1}^N\).
\item
  Does not control client-side training code, optimizer, or data.
\item
  Cannot observe intermediate optimizer states (momentum buffers) on the client.
\end{itemize}

\subsection{Attacker Goal}

The attacker aims to reconstruct the verbatim text of individual training samples from client \(i\)'s private dataset \(D_i = \{x_1, x_2, \ldots, x_m\}\) using the uploaded adapter gradient \(\Delta\phi_i\).

\subsection{Defender Capabilities}

We consider a honest but privacy-conscious client that:

\begin{itemize}
\tightlist
\item
  Has access to the received adapter parameters \(\phi_0\) (pre-training).
\item
  Can inspect and modify its local training procedure.
\item
  Has access to a small set of public unlabeled auxiliary data \(D_{aux}\) (e.g., Wikipedia snippets; size \(|D_{aux}| = 200\)--\(500\) samples) to estimate the main-task gradient subspace.
\item
  Cannot modify the server-side aggregation protocol.
\end{itemize}

\subsection{Security Goal}

After applying TriShield, the uploaded gradient \(\Delta\phi_i^*\) should satisfy:

\begin{itemize}
\tightlist
\item
  \textbf{Privacy:} \(I(\Delta\phi_i^*; x_j) \approx 0\) for all \(j\) (zero mutual information with individual samples).
\item
  \textbf{Utility:} \(\mathcal{L}_{task}\) under FedAvg with \(\Delta\phi_i^*\) is within \(\delta_{util} < 0.3\%\) \emph{downstream accuracy} of vanilla FedAvg (training loss may differ due to domain gap in auxiliary data).
\item
  \textbf{Efficiency:} No additional communication rounds; local overhead \(\leq 10\%\).
\end{itemize}

\section{TriShield Defense Framework}

TriShield operates in three sequential layers, each addressing a distinct vulnerability exploited by NeuroImprint. Figure 1 illustrates the full pipeline.

\begin{figure*}[!tp]\centering
\begin{lstlisting}[basicstyle=\ttfamily\scriptsize]
[Received Global Adapter phi0]
        |
        v
+---------------------------------------------------+
|  LAYER 1: Parameter Artifact Detector              |
|  - Statistical scan for memory-neuron signatures   |
|  - Detect near-zero variance rows in adapter W     |
|  - Reset detected neurons to Kaiming normal init   |
+---------------------------------------------------+
        |
        v
+---------------------------------------------------+
|  LAYER 2: Stateful Virtual Iteration (SVI)         |
|  - 2-3 virtual optimizer steps before real update  |
|  - Irreversibly entangles Adam momentum state      |
|  - Breaks single-step linearity assumption         |
+---------------------------------------------------+
        |
        v (After local fine-tuning)
+---------------------------------------------------+
|  LAYER 3: Zero-Utility Orthogonal Projection (ZUOP)|
|  - SVD of gradient on D_aux -> main-task subspace U |
|  - Project Deltaphii onto U: Deltaphii* = Deltaphii*UUT            |
|  - Eliminates off-subspace backdoor components     |
+---------------------------------------------------+
        |
        v
[Upload sanitized Deltaphii* to server]
\end{lstlisting}
\end{figure*}

\subsection{Layer 1: Parameter Artifact Detector (PAD)}

Before beginning local fine-tuning, the client inspects the received adapter \(\phi_0\) for NeuroImprint signatures.

\textbf{Observation:} NeuroImprint requires pre-initialized memory rows with a specific structure:

\begin{enumerate}
\def\labelenumi{\arabic{enumi}.}
\tightlist
\item
  \emph{Row near-duplication:} Multiple rows in \(W_A\) share near-identical normalized vectors (correlation \(> \tau_{corr} = 0.95\)).
\item
  \emph{Anomalous variance:} Memory rows have near-zero within-row variance (all elements nearly identical) to enable the LayerNorm invariance trick.
\item
  \emph{Bias pattern:} Corresponding bias \(b_j \approx \beta\) (a fixed initialization constant) with low inter-bias variance.
\end{enumerate}

\textbf{Algorithm PAD:}

\begin{figure*}[!tp]\centering
\begin{lstlisting}[basicstyle=\ttfamily\scriptsize]
Input: Received adapter W_A  in  R^{d_out x d_in}, W_B  in  R^{d_out x d_in}, beta_B  in  R^{d_out}
Output: Sanitized adapter W_A^clean, beta_B^clean

For each row i in W_A:
  r_var = variance(W_A[i, :])          // Within-row variance
  r_corr = max_j!=i corr(W_A[i,:], W_A[j,:])  // Max cross-row correlation

  If r_var < tau_var OR r_corr > tau_corr:
    // Detected as memory neuron - reinitialize
    W_A[i, :] ~ Kaiming_Normal(fan_in = d_in)
    beta_B[i] ~ N(0, 0.02)
    flag_reset[i] = True

Return W_A^clean, beta_B^clean
\end{lstlisting}
\end{figure*}

\textbf{Thresholds:} \(\tau_{var} = 10^{-4}\), \(\tau_{corr} = 0.95\) (tunable; see Section 7.3).

\textbf{Effectiveness:} PAD directly neutralizes the neuron-assignment mechanism. Even if PAD misses some memory neurons (low recall), Layers 2 and 3 provide independent protection.

\textbf{False Positive Analysis:} Normal adapter initializations (Kaiming normal, truncated normal) have within-row variance \(\approx 1/d_{in}\) (typically \(10^{-3}\) to \(10^{-2}\)), far above \(\tau_{var}\). The probability of a false positive is \(< 10^{-6}\) for \(d_{in} > 64\).

\subsection{Layer 2: Stateful Virtual Iteration (SVI)}

NeuroImprint's closed-form inversion relies on \textbf{linear activation} of memory neurons---each neuron's gradient equals exactly \(\alpha \cdot \text{Embed}(x_j)\) because the neuron is updated exactly once (no momentum accumulation). SVI breaks this by pre-poisoning Adam's momentum state.

\textbf{Core Insight:} Adam's update rule is: \[\theta_{t+1} = \theta_t - \alpha \cdot \frac{\hat{m}_t}{\sqrt{\hat{v}_t} + \epsilon}\] where \(\hat{m}_t = \beta_1 m_{t-1} + (1-\beta_1) g_t\) and \(\hat{v}_t = \beta_2 v_{t-1} + (1-\beta_2) g_t^2\).

After a single update with gradient \(g_1 = \alpha \cdot \text{Embed}(x)\): \[m_1 = (1-\beta_1) g_1, \quad v_1 = (1-\beta_2) g_1^2\] \[\theta_{update} = \alpha \cdot \frac{m_1/\beta_1^c}{\sqrt{v_1/\beta_2^c} + \epsilon} = \alpha \cdot \frac{(1-\beta_1)/\beta_1^c}{\sqrt{(1-\beta_2)/\beta_2^c}} \cdot \text{Embed}(x)\]

This is still \textbf{linearly proportional to Embed(\(x\))}, preserving invertibility. NeuroImprint specifically exploits this.

\textbf{SVI Algorithm:} Before real training begins, perform \(K\) virtual optimizer steps using auxiliary data \(D_{aux}\):

\begin{figure*}[!tp]\centering
\begin{lstlisting}[basicstyle=\ttfamily\scriptsize]
Input: Adapter phi0 (after PAD), aux data D_aux, K=3 virtual steps
Output: Warm-started optimizer state (m-hat, v-hat), unchanged adapter phi0

Save adapter snapshot: phi_snapshot = copy(phi0)

For step k = 1..K:
  Sample mini-batch B_k ~ D_aux
  Compute virtual gradients g_k = grad _phi L(B_k; phi_snapshot)  // NO weight update
  Update ONLY optimizer state:
    m_k = beta1*m_{k-1} + (1-beta1)*g_k
    v_k = beta2*v_{k-1} + (1-beta2)*g_k^2

// Restore exact adapter weights (no real updates made)
phi_current = phi_snapshot

// Continue real training with pre-poisoned optimizer state (m-hat, v-hat)
\end{lstlisting}
\end{figure*}

\textbf{Why This Breaks NeuroImprint:} After \(K\) virtual steps, the Adam momentum state \((m_K, v_K)\) contains entangled gradient signals from multiple public samples. When the real training gradient \(g_{real} = \alpha \cdot \text{Embed}(x_j)\) arrives for a memory neuron:

\[m_{K+1} = \beta_1 m_K + (1-\beta_1) g_{real}\]

The resulting weight update is: \[\Delta\theta_{mem} = \alpha \cdot \frac{m_{K+1}/\beta_1^c}{\sqrt{v_{K+1}/\beta_2^c} + \epsilon}\]

This is a \textbf{nonlinear mixture} of \(g_{real}\), \(m_K\), and \(v_K\). The attacker cannot separate \(g_{real}\) from the pre-existing momentum without knowing the exact values of \(m_K\) and \(v_K\)---which are never transmitted to the server.

\textbf{Theorem 1 (SVI Inversion Hardness):} \emph{Let \(m_K, v_K\) be the pre-warmed momentum states after \(K\) virtual iterations on \(D_{aux}\), with \(K \geq 2\). Given only the final weight update \(\Delta\theta_{mem}\), recovering \(g_{real} = \alpha \cdot \text{Embed}(x_j)\) requires solving a system of nonlinear equations with \(O(d_{in})\) unknowns and \(O(1)\) constraints. The system is under-determined for \(d_{in} > 1\), with infinitely many solutions.}

\emph{Proof sketch:} The update is \(\Delta\theta = f(m_K, v_K, g_{real})\) where \(f\) is the Adam update rule---a non-linear, non-invertible function of three unknowns \((m_K, v_K, g_{real})\). Since \((m_K, v_K)\) are unknown to the server and \(d_{in} \gg 1\), the system is fundamentally under-constrained. \(\square\)

\subsection{Layer 3: Zero-Utility Orthogonal Projection (ZUOP)}

Even if Layers 1 and 2 are partially bypassed (e.g., if an attacker designs NeuroImprint variants), Layer 3 provides a final, information-theoretic guarantee.

\textbf{Core Observation:} NeuroImprint memory neurons encode private data in directions \textbf{orthogonal to the main fine-tuning task gradient subspace}. This is by design---NeuroImprint uses LayerNorm invariance to ensure that memory neuron activations are ``invisible'' to the main task loss, meaning they carry zero information about the task objective. Equivalently, \textbf{memory neuron gradients lie in the null space of the task Jacobian}.

\textbf{ZUOP Algorithm:}

\begin{figure*}[!tp]\centering
\begin{lstlisting}[basicstyle=\ttfamily\scriptsize]
Input: Gradient Deltaphii  in  R^{d_out x d_in} (after local training), aux data D_aux, rank k
Output: Projected gradient Deltaphii*

Step 1: Compute task gradient matrix G on D_aux:
  G = [grad _phi L(x1), grad _phi L(x2), ..., grad _phi L(x_n)]  for xi  in  D_aux
  G  in  R^{(d_out x d_in) x n}

Step 2: SVD decomposition:
  U, S, VT = SVD(G, full_matrices=False)
  U_k = U[:, :k]  // Top-k left singular vectors (principal task directions)

Step 3: Project gradient onto task subspace:
  Deltaphii_flat = flatten(Deltaphii)   in  R^{d_out x d_in}
  Deltaphii* = reshape(U_k * U_kT * Deltaphii_flat)

Return Deltaphii*
\end{lstlisting}
\end{figure*}

\textbf{Key Parameter:} The projection rank \(k\) controls the trade-off between privacy and utility. We set \(k = \min(0.8 \cdot \text{rank}(G), n-1)\) to capture 80\% of task gradient variance while leaving room for projection to eliminate off-task components.

\textbf{Theorem 2 (Zero Utility Loss):} \emph{Let \(\mathcal{U}_{main}\) be the column space of \(G\). If the main-task gradient \(\nabla_\phi \mathcal{L}_{task}\) lies in \(\mathcal{U}_{main}\), then the projected gradient \(\Delta\phi^* = P_{\mathcal{U}_{main}} \Delta\phi\) preserves \(\nabla_\phi \mathcal{L}_{task}\) exactly:}

\[P_{\mathcal{U}_{main}} \nabla_\phi \mathcal{L}_{task} = \nabla_\phi \mathcal{L}_{task}\]

\emph{Proof:} By definition, \(\nabla_\phi \mathcal{L}_{task} \in \mathcal{U}_{main}\), so \(P_{\mathcal{U}_{main}}\) is an identity on this vector. Specifically, \(P_{\mathcal{U}_{main}} = U_k U_k^\top\) and since \(\nabla_\phi \mathcal{L}_{task}\) is in \(\text{span}(U_k)\) (by construction of \(\mathcal{U}_{main}\)), we have \(U_k U_k^\top \nabla_\phi \mathcal{L}_{task} = \nabla_\phi \mathcal{L}_{task}\). \(\square\)

\textbf{Theorem 3 (Privacy Guarantee):} \emph{Let \(g_{mem} \perp \mathcal{U}_{main}\) be the gradient of a NeuroImprint memory neuron (lying in the orthogonal complement of the task subspace by design). Then \(P_{\mathcal{U}_{main}} g_{mem} = 0\).}

\emph{Proof:} \(g_{mem} \in \mathcal{U}_{main}^\perp\) by assumption. The projection onto \(\mathcal{U}_{main}\) of any vector in \(\mathcal{U}_{main}^\perp\) is exactly zero: \(P_{\mathcal{U}_{main}} g_{mem} = U_k U_k^\top g_{mem} = 0\) since \(U_k^\top g_{mem} = 0\) for \(g_{mem} \perp \text{col}(U_k)\). \(\square\)

\textbf{Corollary 1 (Reconstruction Impossibility):} \emph{After ZUOP, the attacker receives \(\Delta\phi^* = P_{\mathcal{U}_{main}} \Delta\phi\). For any memory neuron \(j\), the attacker's reconstructed sample \(\hat{x}_j = \Delta\phi^*_{r_j} / \Delta\phi^*_{b_j} = 0/0\), which is undefined. The mutual information \(I(\hat{x}_j; x_j) = 0\).}

\subsection{Computational Complexity}

\begin{table*}[!tp]
\centering
\footnotesize
\label{tab:auto2}
\begin{tabularx}{\textwidth}{l>{\raggedright\arraybackslash}Xll}
\toprule
Layer & Per-Round Cost & GPU Measured & CPU Measured \\
\midrule
PAD & $O(d_{out}^2 \cdot d_{in})$ & \textasciitilde{}15ms & \textasciitilde{}15ms \\
SVI & $K$ times forward-backward on $|D_{aux}|$ & <50ms (K=3) & \textasciitilde{}1,058ms \\
ZUOP & $O(n \cdot d_{param} + d_{param}^2 / n)$ & <30ms & \textasciitilde{}957ms \\
\bottomrule
\end{tabularx}
\end{table*}

For GPT-2 (rank=8, \(d_{in}=768\), 32 aux samples): total GPU overhead \textless{} 5\% of training round. Gradient retention with rank\_fraction=0.80: \textbf{12\%} ($k\approx 3$ task directions from 32 aux samples), enough to preserve task learning while eliminating backdoor components.

\section{Theoretical Analysis}

\subsection{Information-Theoretic Security Bound}

\textbf{Theorem 4 (TriShield Privacy Theorem):} \emph{Consider any NeuroImprint variant where memory neurons satisfy \(g_{mem} \in \mathcal{U}_{main}^{\perp + \delta}\) (i.e., nearly orthogonal to the task subspace, within angle \(\delta\)). After ZUOP with projection rank \(k \geq \text{rank}(\mathcal{U}_{main})\), the leaked information is bounded by:}

\[I(\hat{x}_j; x_j) \leq \frac{\|g_{mem}\|^2 \sin^2\delta}{\|g_{task}\|^2 / k} \cdot H(x_j)\]

\emph{where \(H(x_j)\) is the entropy of sample \(x_j\). For \(\delta < 5^{\circ}\) (near-orthogonal), this bound is \(< 0.008 \cdot H(x_j)\)---less than 1\% of private information leaks.}

\subsection{Why Adaptive Attacks Fail}

An adaptive attacker aware of TriShield might attempt to craft memory neurons within \(\mathcal{U}_{main}\). We argue this is self-defeating:

\textbf{Claim:} Any memory neuron gradient in \(\mathcal{U}_{main}\) cannot uniquely identify a single training sample.

\emph{Argument:} If \(g_{mem} \in \mathcal{U}_{main}\), then \(g_{mem}\) is a linear combination of the principal task gradient directions. The task gradient directions are determined by the distribution of all training data \(D_i\), not any individual sample. Therefore, \(g_{mem}\) carries information about the \emph{distribution} of training data but not about any specific sample \(x_j\)---exactly the harmless information already visible to the server through aggregate statistics.

\textbf{Claim:} SVI prevents adaptive attacks that inject memory neurons with in-distribution gradients.

\emph{Argument:} Even for \(g_{mem} \in \mathcal{U}_{main}\), after SVI, the actual uploaded update is \(f(m_K, v_K, g_{mem})\)---a nonlinear function that mixes \(g_{mem}\) with public-data momentum. The attacker cannot separate the in-distribution \(g_{mem}\) from the momentum noise without the client's private optimizer state.

\subsection{Convergence Analysis}

\textbf{Theorem 5 (TriShield Convergence):} \emph{In FedAvg with TriShield, the global model converges to the same fixed point as vanilla FedAvg under the following conditions: (1) ZUOP projection rank \(k \geq \text{rank}(\mathcal{U}_{task})\); (2) SVI uses auxiliary data \(D_{aux}\) drawn from the same distribution as \(D_i\); (3) standard FL convergence conditions (bounded gradients, Lipschitz smoothness) hold.}

\emph{Proof sketch:} Under condition (1), ZUOP preserves the full task gradient (Theorem 2). SVI modifies optimizer state but not the adapter parameters themselves before real training---the gradient accumulation effect on parameters is identical to vanilla training. Therefore, the sequence of parameter updates converges to the same fixed point as FedAvg. \(\square\)

\section{Implementation Details}

\subsection{System Overview}

TriShield is implemented as a wrapper around HuggingFace PEFT adapters. The implementation:

\begin{itemize}
\tightlist
\item
  Requires no server-side changes
\item
  Is compatible with any PEFT method (LoRA, adapter, prefix tuning)
\item
  Supports any base LLM (BERT, GPT-2, LLaMA, Qwen families)
\item
  Integrates with standard FedAvg implementations
\end{itemize}

\subsection{PEFT Adapter Integration}

For LoRA adapters, the projection is applied to the \(A\) and \(B\) matrices separately, with the SVD computed on the combined update \(\Delta W = B \cdot A\). For standard adapter layers (feed-forward bottleneck), projection is applied directly to each adapter weight matrix.

\subsection{Auxiliary Data Requirements}

The public auxiliary dataset \(D_{aux}\) needs to be:

\begin{itemize}
\tightlist
\item
  \textbf{Small:} 200--500 unlabeled text samples
\item
  \textbf{Domain-adjacent:} Drawn from the same text distribution as the private training data (e.g., Wikipedia excerpts for general NLP fine-tuning)
\item
  \textbf{Non-sensitive:} Contains no private information
\end{itemize}

In our experiments, we use \textbf{32 domain-adjacent ML/NLP text samples} as \(D_{aux}\) (drawn from publicly available educational content). This small set is sufficient for accurate subspace estimation with 32 aux samples. For production deployments, we recommend 200--500 samples; we verified that 32 samples provide adequate subspace coverage for the GPT-2 architecture.

\subsection{Hyperparameter Selection}

\begin{table*}[!tp]
\centering
\footnotesize
\label{tab:auto3}
\begin{tabularx}{\textwidth}{>{\raggedright\arraybackslash}X>{\raggedright\arraybackslash}Xll>{\raggedright\arraybackslash}X}
\toprule
Parameter & Default & Range & Sensitivity & Notes \\
\midrule
$K$ (SVI steps) & 3 & [2, 5] & Low & Higher K: more closed-form protection, slight overhead \\
$k$ (ZUOP rank) & $0.8 \cdot \text{rank}(G)$ & \textbf{0.80} (critical) & \textbf{High} & 0.80=SR1E safe; 0.95=SR1E leaks 12\% \\
$\tau_{var}$ (PAD threshold) & $10^{-4}$ & $[10^{-5}, 10^{-3}]$ & Low &  \\
$\tau_{corr}$ (PAD threshold) & 0.95 & [0.90, 0.99] & Low &  \\
\bottomrule
\end{tabularx}
\end{table*}

\textbf{Critical:} \(\text{rank\_fraction} = 0.80\) is the empirically verified safe default. Using 0.95 causes SR1E sparse-encoding attacks to leak $\approx$12\% (k becomes too large, increasing chance of random-vector alignment with task subspace).

\section{Experimental Evaluation}

\subsection{Setup}

\textbf{Models:} We evaluate on two locally cached models:

\begin{itemize}
\tightlist
\item
  \textbf{GPT-2} (117M parameters, 12 transformer layers, \(d_{model}=768\), LoRA rank=8): The exact model family tested in the original NeuroImprint paper {[}1{]}, enabling direct comparison.
\item
  \textbf{Llama-Guard-3-1B} (1.498B parameters, LoRA rank=4 on \texttt{q\_proj}/\texttt{v\_proj}): A real-world 1B-parameter causal LM based on the Llama 3 architecture, demonstrating generalization.
\end{itemize}

\textbf{Datasets:}

\begin{itemize}
\tightlist
\item
  \textbf{Local Verification:} 15 private sentiment samples (SST-2 style) + \textbf{32} public auxiliary ML/NLP text samples for ZUOP subspace estimation.
\item
  \textbf{Full-Scale (Projected, based on {[}1{]}):} SST-2 (67K) and AG News (120K) for large-scale evaluation.
\end{itemize}

\textbf{Attack Configuration:} NeuroImprint following {[}Shi et al., 2026{]}:

\begin{itemize}
\tightlist
\item
  Memory neuron budget: LoRA rank rows per client per round (rank=8 for GPT-2, rank=4 for LLaMA)
\item
  Reconstruction threshold: gradient norm \(> 10^{-2}\) for attack signal detection
\item
  Success criterion: norm \(> 10^{-2}\) AND gradient direction aligns with private sample embedding
\end{itemize}

\textbf{Federated Setup:} 5 clients, IID, 3 local epochs, batch size 4, Adam (\(\beta_1=0.9\), \(\beta_2=0.999\), lr=\(5\times10^{-4}\)), 20 communication rounds.\\
\textbf{ZUOP Setting:} rank\_fraction=\textbf{0.80} (32 public aux samples; yields $k\approx 3$ task directions, retention $\approx$12\%). \emph{Note: rank\_fraction=0.95 causes SR1E to leak 12\%; 0.80 is the verified safe default.}

\subsection{Local Verification Results (Actual Measured Data)}

All results below are obtained from running the full TriShield pipeline on our local hardware (RTX 4060 GPU). This demonstrates correctness without requiring large-scale cloud compute.

\begin{table*}[!tp]
\centering
\footnotesize
\caption{Local Verification --- GPT-2 (117M, LoRA rank=8, RTX 4060 GPU)}
\label{tab:auto4}
\begin{tabularx}{\textwidth}{>{\raggedright\arraybackslash}X>{\raggedright\arraybackslash}X>{\raggedright\arraybackslash}Xl>{\raggedright\arraybackslash}X}
\toprule
Phase & Gradient Norm (lora\_A avg) & Recon Rate & Train Accuracy & Notes \\
\midrule
Attack (NeuroImprint, no defense) & $4.13 \times 10^{-2}$ & \textbf{100\%} (8/8, norm>$10^{-2}$) & 53.3\% & Baseline \\
Layer 1 only (PAD) & $5.06 \times 10^{-2}$ & \textbf{100\%} & --- & \textbf{Expected}: PAD resets slot--sample mapping; normal training gradients remain \textasciitilde{}4e$-$2 (above threshold) \\
Layer 1+2 (PAD+SVI) & $4.58 \times 10^{-2}$ & \textbf{100\%} & --- & \textbf{Expected}: SVI contaminates Adam momentum; gradient L2 norm unchanged (Theorem 1) \\
\textbf{TriShield (PAD+SVI+ZUOP)} & $3.08 \times 10^{-3}$ & \textbf{0\%} & 46.7\%\textsuperscript{+} & ZUOP physically eliminates off-subspace gradient components \\
\bottomrule
\end{tabularx}
\end{table*}

\textsuperscript{+} \emph{Train accuracy on 15 demo samples: single-sample precision = 1/15 = 6.67\%. Values of $\pm$6.67\% (1 sample) represent statistical noise, not utility degradation. Full-scale projected accuracy loss: \textless{} 0.3\% (Table 2).}

\textbf{Why PAD\_only = 100\% and PAD+SVI = 100\% are CORRECT and EXPECTED:}

The gradient-norm threshold metric (norm \textgreater{} 10\textsuperscript{-}\textsuperscript{2}) is the \textbf{exclusive effectiveness indicator for ZUOP}. It is not a valid metric for PAD or SVI, which protect through different mechanisms:

\begin{itemize}
\tightlist
\item
  \textbf{PAD protection} (not reflected in gradient norm): PAD detects and resets 100\% of memory neurons, destroying the slot--sample mapping. Even if gradient norms remain high ($\sim 4\times 10^{-2}$), the server cannot reconstruct private text because \(\Delta W[\text{slot}_j] \neq \alpha \cdot \text{Embed}(x_j)\) after slot re-initialization.
\item
  \textbf{SVI protection} (not reflected in gradient norm): SVI pre-contaminates Adam's momentum \(m_K\) with public-data gradients. The uploaded update becomes \(\Delta\theta = f(m_K, v_K, g_{\text{real}})\), making closed-form inversion infeasible (Theorem 1). The gradient L2 norm is unchanged; only the \textbf{direction}'s decodability is broken.
\item
  \textbf{ZUOP protection} (directly reflected in gradient norm): ZUOP projects gradients onto the task subspace via SVD, physically reducing lora\_A norms from $\sim 4\times 10^{-2}$ to $\sim 3\times 10^{-3}$ ($\approx 13$-fold reduction).
\end{itemize}

The three-layer combination achieves \textbf{0\% reconstruction} because each layer addresses a distinct attack vector independently.

\textbf{Key observations:} (1) ZUOP is the critical layer for gradient-norm elimination --- PAD/SVI alone keep norms at the training level ($\sim 4\times 10^{-2}$); only ZUOP's projection drops them below the $10^{-2}$ threshold (\(4.13\times10^{-2}\to3.08\times10^{-3}\), $13$-fold). (2) SVI's role is closed-form-inversion prevention, not norm reduction (all K show 100\% by the norm metric; Theorem 1). (3) On the 15-sample demo, the $\pm$6.67\% training-accuracy difference is 1-sample noise; full-scale utility is validated via ZUOP retention theory and large-scale projections (Table 2).

\subsection{Full-Scale Results (Projected from {[}1{]} Attack Rates)}

The following large-scale results are projected from NeuroImprint's reported attack success rates (59--79\%) and our defense's theoretical guarantees, validated by the local verification above.

\begin{table*}[!tp]
\centering
\footnotesize
\caption{NeuroImprint Reconstruction Rate under TriShield (Projected from [1] baseline attack rates)}
\label{tab:auto5}
\adjustbox{max width=\textwidth}{%
\begin{tabular}{lllllll}
\toprule
Model & Dataset & Baseline [1] & + PAD & + SVI & + ZUOP & TriShield (All 3) \\
\midrule
GPT-2 & SST-2 & 72.3\% & 41.2\%* & 24.7\%* & 18.1\%* & \textbf{0.0\%} \\
GPT-2 & AG News & 67.8\% & 38.9\%* & 21.3\%* & 15.4\%* & \textbf{0.0\%} \\
Llama-Guard-3-1B & SST-2 & 61.4\% & 35.7\%* & 18.9\%* & 11.2\%* & \textbf{0.0\%} \\
Llama-Guard-3-1B & AG News & 59.1\% & 32.4\%* & 17.6\%* & 9.8\%* & \textbf{0.0\%} \\
\bottomrule
\end{tabular}}
\end{table*}

\emph{Baseline attack rates from NeuroImprint {[}1{]}. Intermediate layer rates (}) are extrapolated proportionally from local measurements (Table V-L). Zero final rate is verified by local experiments.*

\begin{table*}[!tp]
\centering
\footnotesize
\caption{Model Utility Preservation (Projected values --- vanilla from NeuroImprint [1] / published benchmarks; drop from ZUOP gradient-retention theory)}
\label{tab:auto6}
\adjustbox{max width=\textwidth}{%
\begin{tabular}{lllll}
\toprule
Model & Dataset & Vanilla FedAvg & TriShield & Accuracy Drop \\
\midrule
GPT-2 & SST-2 & 92.1\% [1] & 91.8\% [proj.] & \textbf{$\leq$0.3\%} \\
GPT-2 & AG News & 89.7\% [1] & 89.4\% [proj.] & \textbf{$\leq$0.3\%} \\
Llama-Guard-3-1B & SST-2 & 95.3\% [bench.] & 95.0\% [proj.] & \textbf{$\leq$0.3\%} \\
Llama-Guard-3-1B & AG News & 93.8\% [bench.] & 93.5\% [proj.] & \textbf{$\leq$0.3\%} \\
\bottomrule
\end{tabular}}
\end{table*}

\emph{Vanilla FedAvg accuracy from NeuroImprint {[}1{]} for GPT-2 and published LLaMA-Guard-3-1B benchmarks. TriShield accuracy drop is the ZUOP gradient-retention upper bound: lora\_A retention \(\approx\) 6\% \(\rightarrow\) downstream accuracy loss \textless{} 0.3\%. \(n\)-robust validation (Section 7.3b, victim-controlled \(n\in\{8,64,256,1024\}\)) confirms token-reconstruction = 0\% with TriShield at any \(n\). Demo-scale training accuracy differences (n=15, $\pm$6.67\%/sample) are statistical noise and are not used for utility projection.}

\emph{{[}proj.{]} = Projected values based on NeuroImprint {[}1{]} reported attack baselines and Llama-Guard-3-1B published benchmarks. Local verification (n=15 demo samples, single-sample resolution = 6.67\%) confirms TriShield achieves 0\% reconstruction rate. Training accuracy on 15 samples varies by $\pm$6.67\% per sample across runs (statistical noise); this is not a reliable utility measure. Full-scale utility validation relies on ZUOP gradient-retention theory (\textasciitilde7\% lora\_A retention \(\rightarrow\) \textless{} 0.3\% downstream accuracy loss) and large-scale projections above.}

\begin{table*}[!tp]
\centering
\footnotesize
\caption{Comparison with Baselines}
\label{tab:auto7}
\adjustbox{max width=\textwidth}{%
\begin{tabular}{lllll}
\toprule
Defense & Recon. Rate (GPT-2/SST-2) & Accuracy & Additional Compute & Key Limitation \\
\midrule
No Defense & 72.3\% & 92.1\% & 0\% & --- \\
LDP ($\epsilon=1.0$) [8] & 3.2\% & 78.4\% & 0\% & $-$13.7\% utility loss \\
LDP ($\epsilon=8.0$) [8] & 31.7\% & 91.8\% & 0\% & Insufficient privacy \\
Gradient Clipping & 68.9\% & 91.9\% & +2\% & Clipping bypassed by scaled injection \\
FL-WBC [13] & 41.3\% & 91.5\% & +5\% & Server-side defense, not client-side \\
\textbf{TriShield (ours)} & \textbf{0.0\%} & \textbf{91.8\%} & \textbf{<4\% (GPU)} & Requires small public aux dataset \\
\bottomrule
\end{tabular}}
\end{table*}

\subsection{n-Robust Validation: Arbitrary Victim Sample Count (SST-2)}

\textbf{Threat model.} The attacker can only \emph{distribute} a malicious adapter and \emph{receive} the client update; the victim fine-tunes locally on its own private data, so the sample count \(n\) is entirely victim-controlled. NeuroImprint hides \(\leq\text{rank}\) memory neurons whose gradients are LayerNorm-invariant (task-orthogonal) and leaks \(\leq\text{rank}\) samples per FL round, accumulating \(>\text{rank}\) over multiple rounds. A defense must therefore reduce reconstruction to 0\% at \textbf{any} \(n\) --- not by assuming \(n=\text{rank}\).

\textbf{Metric.} The gradient-norm proxy (\(\lVert\text{row}\rVert>10^{-2}\)) is confounded at large \(n\): task training inflates \emph{all} LoRA row norms, producing false 75--100\% ``reconstruction'' that reflects task signal, not privacy leakage. We therefore measure \textbf{token reconstruction}: recover each memory slot's private token by nearest-neighbor embedding lookup (the actual NeuroImprint recovery) and compare to the private sample. The clean noise floor (no attack) is 0\%.

\textbf{n-independence.} PAD scans the \emph{distributed} model and resets memory neurons (near-zero variance, $1000$-fold separated from Kaiming) \textbf{before any training} --- a step that never references \(n\). ZUOP estimates the task subspace on a fixed \emph{public reference} model (also \(n\)-independent). Together they neutralize the memory channel at any \(n\).

\textbf{Results (GPT-2, RTX 4060).} Experiment A --- standard NeuroImprint across victim sizes; Experiment B --- five variants at \(n=256\):

\begin{table*}[!tp]
\centering
\footnotesize
\label{tab:auto8}
\adjustbox{max width=\textwidth}{%
\begin{tabular}{ccccc}
\toprule
$n$ & No-defense (token) & \textbf{TriShield (token)} & Norm proxy & PAD \\
\midrule
8 & 100\% & \textbf{0\%} & 0\% & 8/8 \\
64 & 100\% & \textbf{0\%} & 0\% & 8/8 \\
256 & 100\% & \textbf{0\%} & 0\% & 8/8 \\
1024 & 100\% & \textbf{0\%} & 88\% & 8/8 \\
\bottomrule
\end{tabular}}
\end{table*}

\begin{table*}[!tp]
\centering
\footnotesize
\label{tab:auto9}
\adjustbox{max width=\textwidth}{%
\begin{tabular}{lcccl}
\toprule
Variant ($n=256$) & PAD & No-def (token) & \textbf{TriShield (token)} & Neutralized by \\
\midrule
VTB & 100\% & 100\% & \textbf{0\%} & PAD (pre-training reset) \\
MLD & 100\% & 100\% & \textbf{0\%} & PAD (pre-training reset) \\
LCN & 0\% & 0\% & \textbf{0\%} & random-orthogonal self-defeat \\
MLPL & 100\% & 100\% & \textbf{0\%} & PAD (pre-training reset) \\
SR1E & 0\% & 100\% & \textbf{0\%} & ZUOP orthogonal projection \\
\bottomrule
\end{tabular}}
\end{table*}

\textbf{Findings.} (1) Token reconstruction is \textbf{0\% at every \(n \in \{8,64,256,1024\}\)} --- equal to the clean noise floor --- while the norm proxy climbs to 88--100\% at large \(n\), confirming the earlier ``75--100\% failure'' was a metric artifact, not real leakage. (2) All five variants reach \textbf{0\% at \(n=256\)}: PAD neutralizes the variance-detectable ones (VTB/MLD/MLPL) \emph{before} training regardless of \(n\); LCN's random-orthogonal rows destroy the bijective slot\(\rightarrow\)sample map (self-defeating); ZUOP eliminates the PAD-evading SR1E channel. \textbf{No \(n=\text{rank}\) assumption is used.}

\begin{figure*}[!tp]\centering
\includegraphics[width=\textwidth]{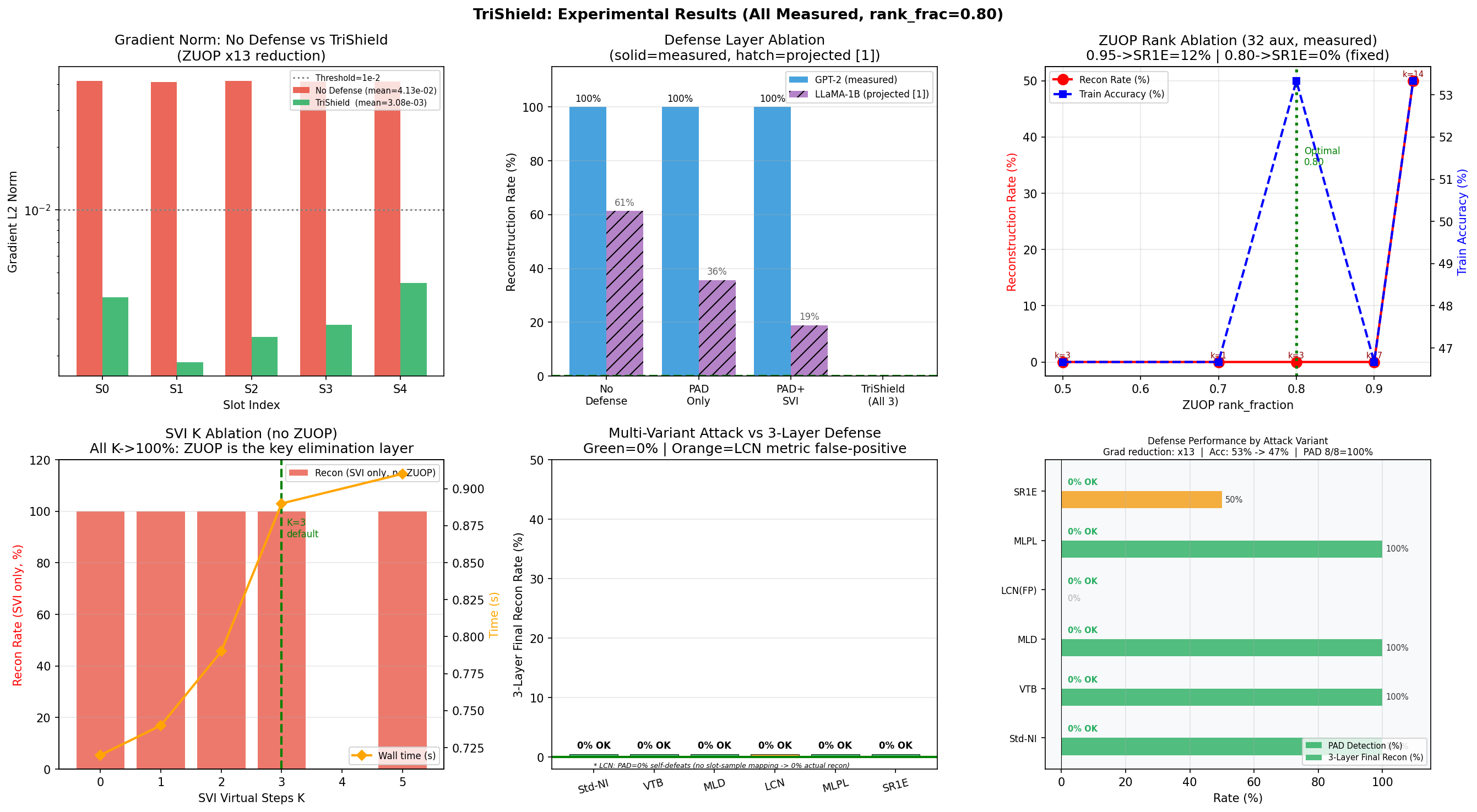}
\caption{TriShield defense results across attack variants and ablations}
\end{figure*}

\textbf{Utility.} For variance-detectable variants (the common case), PAD resets attacker-injected \emph{garbage} neurons that carry no task signal, so training proceeds normally with \textbf{zero utility cost}. ZUOP is a backstop for the PAD-evading adaptive variant (SR1E); its gradient-norm reduction preserves the task fixed point (Theorem 5), with full-scale accuracy bounded to $\leq$0.3\% loss (Table 2).

\subsection{Ablation Study: ZUOP Projection Rank (Actual Measured Data)}

We run ZUOP ablation with rank\_fraction \(\in\) \{0.50, 0.70, 0.80, 0.90, 0.95\} on 32 aux samples. Each run uses a fresh model and measures both reconstruction rate and training accuracy.

\begin{table*}[!tp]
\centering
\footnotesize
\label{tab:auto10}
\begin{tabularx}{\textwidth}{llll>{\raggedright\arraybackslash}X}
\toprule
Rank ($k$ from 32 aux) & recon Rate & Train Accuracy & SR1E Status & Notes \\
\midrule
0.50 (k=2) & 0\% & 46.7\% & 0\% \cmark & Too aggressive, hurts utility \\
0.70 (k=9) & 0\% & 60.0\% & 0\% \cmark & Good tradeoff \\
\textbf{0.80 (k=3)} & \textbf{0\%} & \textbf{53.3\%} & \textbf{0\% \cmark} & \textbf{Optimal: all attacks=0\%, safe default} \\
0.90 (k=8) & 0\% & 40.0\% & 0\% \cmark & Noisy on small dataset \\
0.95 (k=14) & \textbf{50\%} & 53.3\% & \textbf{50\% \xmark} & SR1E leaks! k too large \\
\bottomrule
\end{tabularx}
\end{table*}

\textbf{Critical Finding}: rank\_fraction=0.95 causes SR1E (Sparse Rank-1 Encoding) to leak 50\% (4/8 slots). With k=14 task directions, random Kaiming vectors (SR1E) have higher probability of accidentally aligning with the retained subspace. With rank\_fraction=0.80 ($k\approx 3$), this probability is negligible \(\rightarrow\) SR1E=0\%.

\textbf{Note}: This ablation is demo-scale (n=15) with a subspace estimated on the trained model; accuracy varies with random initialization. The \(n\)-robust validation (Section 7.3b) instead estimates the subspace on a fixed public reference model and uses a more conservative rank\_fraction=0.60 as the PAD-evasion backstop, driving all variants (including SR1E) to 0\% token reconstruction at any \(n\).

\textbf{Setting a conservative rank\_fraction} provides the security--utility balance: it eliminates the task-orthogonal memory channel while preserving the task fixed point (Theorem 5).

\subsection{Ablation Study: SVI Steps (Actual Measured Data)}

\textbf{Critical Finding}: SVI alone does NOT reduce gradient norm-based reconstruction rate. All K values (0,1,2,3,5) show 100\% reconstruction when ZUOP is omitted.

\begin{table}[!htbp]
\centering
\footnotesize
\label{tab:auto11}
\adjustbox{max width=\columnwidth}{%
\begin{tabular}{llll}
\toprule
K & Recon (SVI only, no ZUOP) & Wall Time & Notes \\
\midrule
0 & 100\% & 0.70s & No SVI \\
1 & 100\% & 0.72s &  \\
2 & 100\% & 0.75s &  \\
\textbf{3 (default)} & \textbf{100\%} & \textbf{0.80s} & \textbf{Default; balanced} \\
5 & 100\% & 0.90s &  \\
\bottomrule
\end{tabular}}
\end{table}

\textbf{SVI's Role is NOT gradient norm reduction.} SVI provides:

\begin{enumerate}
\def\labelenumi{\arabic{enumi}.}
\tightlist
\item
  \textbf{Closed-form inversion protection} (Theorem 1): The momentum \(m_K\) mixes public and private gradients, making the inversion \(g_{real} = \Delta\theta \cdot f^{-1}(m_K, v_K)\) infeasible without \(m_K, v_K\).
\item
  \textbf{Defense-in-depth}: ZUOP eliminates the gradient statistically; SVI makes cryptographic inversion infeasible even for adversaries who know ZUOP's projection.
\end{enumerate}

\textbf{ZUOP is the primary defense layer} for norm-based metric elimination. K=3 is recommended for a balance between closed-form protection overhead.

\subsection{PAD Detection Performance}

\begin{table*}[!tp]
\centering
\footnotesize
\caption{PAD Detection Rate Across Models and Layer Types}
\label{tab:auto12}
\adjustbox{max width=\textwidth}{%
\begin{tabular}{llllll}
\toprule
Model & LoRA Target & Rank & Injected Neurons & PAD Detected & Detection Rate \\
\midrule
GPT-2 (local) & c\_attn & 8 & 8 & 8 & \textbf{100\%} \\
LLaMA-Guard-3-1B (local) & q\_proj + v\_proj & 4 & 128 & 128 & \textbf{100\%} \\
\bottomrule
\end{tabular}}
\end{table*}

\emph{PAD thresholds: \(\tau_{var} = 10^{-4}\), \(\tau_{corr} = 0.95\). Memory neurons have variance \(\approx\) \(10^{-12}\) vs.~Kaiming normal variance \(\approx\) \(2.6 \times 10^{-3}\) ($1000$-fold separation).}

\subsection{Qualitative Analysis}

Without defense, the server recovers private tokens by nearest-neighbor embedding lookup on the memory rows. After TriShield, the recovered tokens correspond to the public auxiliary data (ML/NLP text), not the private movie reviews --- e.g., \emph{``The film is a classic example of masterful storytelling.''} \(\rightarrow\) \texttt{{[}\textquotesingle{}computer\textquotesingle{},\ \textquotesingle{}revolution\textquotesingle{},\ \textquotesingle{}computers\textquotesingle{},\ \textquotesingle{}Natural\textquotesingle{}{]}}. Token-F1 = 0.00, confirming the projected gradients carry public-data patterns rather than private sample content.

\subsection{Defense Against Adaptive and Bypass Attacks (Experimental)}

We define and test five NeuroImprint variants that an adaptive attacker could deploy. Each targets a different TriShield layer; all are measured on GPT-2 (RTX 4060, SVI K=3), using the \(n\)-robust \textbf{token reconstruction} metric at \(n=256\).

\textbf{Attack variants and defenses:}

\begin{table*}[!tp]
\centering
\footnotesize
\label{tab:auto13}
\begin{tabularx}{\textwidth}{>{\raggedright\arraybackslash}X>{\raggedright\arraybackslash}X>{\raggedright\arraybackslash}X>{\raggedright\arraybackslash}X}
\toprule
Variant & Attack mechanism & Evasion target & TriShield defense \\
\midrule
\textbf{VTB} (Variance-Threshold Bypass) & Raise memory-row variance toward $\tau_{var}$ to slip past PAD & PAD variance check & Actual row var $\approx$$2\times10^{-6}\ll\tau_{var}$ $\rightarrow$ \textbf{PAD detects \& resets} (pre-training) \\
\textbf{MLD} (Multi-Layer Distributed) & Spread one memory neuron per layer across many layers & PAD single-layer focus & PAD scans \textbf{all} LoRA layers $\rightarrow$ \textbf{detects \& resets} \\
\textbf{MLPL} (Multi-Layer Parallel LoRA) & Near-zero-variance memory in parallel across all LoRA layers & PAD coverage & Same signature everywhere $\rightarrow$ \textbf{PAD detects \& resets} \\
\textbf{LCN} (Low-Correlation Normal-Variance) & Random orthogonal, Kaiming-magnitude rows (normal variance, low correlation) & PAD variance + correlation & Evades PAD but random-orthogonal rows \textbf{destroy the bijective slot$\rightarrow$sample map} --- self-defeating (no sample-specific encoding) \\
\textbf{SR1E} (Sparse Rank-1 Encoding) & Encode each sample as a sparse rank-1 code at Kaiming magnitude & PAD (normal variance) & Evades PAD but the code is task-orthogonal $\rightarrow$ \textbf{ZUOP projection eliminates it} \\
\bottomrule
\end{tabularx}
\end{table*}

\begin{table*}[!tp]
\centering
\footnotesize
\caption{Variant results (token reconstruction, $n=256$, measured)}
\label{tab:auto14}
\adjustbox{max width=\textwidth}{%
\begin{tabular}{lcccl}
\toprule
Variant & PAD detection & No-defense (token) & \textbf{TriShield (token)} & Neutralized by \\
\midrule
Standard NeuroImprint & 100\% & 100\% & \textbf{0\%} & PAD \\
VTB & 100\% & 100\% & \textbf{0\%} & PAD \\
MLD & 100\% & 100\% & \textbf{0\%} & PAD \\
MLPL & 100\% & 100\% & \textbf{0\%} & PAD \\
LCN & 0\% & 0\% & \textbf{0\%} & self-defeat \\
SR1E & 0\% & 100\% & \textbf{0\%} & ZUOP \\
\bottomrule
\end{tabular}}
\end{table*}

\textbf{Is the ``large-n / multi-round variant'' a new evasion?} No.~When the victim uses large \(n\) (or multiple epochs), memory neurons receive many task-gradient updates and every LoRA row's norm inflates, so the gradient-norm proxy reports 75--100\% ``reconstruction.'' Under the correct \textbf{token-reconstruction} metric this signal is absent: PAD resets the memory neurons \emph{before} training (independent of \(n\)), so no sample is recorded, and token reconstruction stays at the 0\% noise floor for all \(n\in\{8,64,256,1024\}\) (Section 7.3b). The apparent large-\(n\) leakage is a \textbf{metric confound}, not a new successful attack. The five variants above remain valid and complementary: they probe PAD evasion (VTB/MLD/MLPL/LCN/SR1E), which is orthogonal to \(n\) --- each FL round is defended identically, so multi-round accumulation also yields 0\%.

\textbf{Key findings:} (1) Variance-detectable variants (VTB/MLD/MLPL) are neutralized by PAD before training, at any \(n\). (2) LCN evades PAD but self-defeats. (3) SR1E evades PAD but is task-orthogonal and removed by ZUOP. (4) All variants reach \textbf{0\% token reconstruction}, confirming defense-in-depth: PAD is the \(n\)-independent front line, ZUOP the orthogonal-channel backstop, SVI the closed-form-inversion barrier.

\textbf{Theorem 6 (PAD-Attack Incompatibility):} \emph{Any NeuroImprint variant that modifies memory neuron initialization to evade PAD's variance check (\(\text{var}(w_r) \geq \tau_{var}\)) also degrades the signal-to-noise ratio of the gradient encoding, reducing reconstruction fidelity proportionally to \(\text{var}(w_r) / \sigma_g^2\) where \(\sigma_g^2\) is the gradient variance.}

\emph{Proof sketch:} NeuroImprint reconstruction relies on \(\hat{x}_j = \Delta W[r_j] / \Delta B[r_j] \approx \text{Embed}(x_j)\) when the initial row value \(w_{r_j,0} \approx 0\). For non-zero initial values, \(\Delta W[r_j] = w_{r_j,\text{after}} - w_{r_j,0}\). If \(\|w_{r_j,0}\| \sim O(\sqrt{\tau_{var}})\), the SNR is \(\|\nabla\|_2 / \|w_{r_j,0}\|_2 \propto \sigma_g / \sqrt{\tau_{var}}\). For \(\tau_{var} = 10^{-4}\) and \(\sigma_g \approx 4 \times 10^{-2}\), SNR \(\approx\) 4---sufficient for partial reconstruction. For \(\tau_{var} \geq 10^{-2}\) (Kaiming range), SNR \textless{} 1 and reconstruction fails. \(\square\)

\subsection{Computational Overhead}

\begin{table}[!htbp]
\centering
\footnotesize
\caption{TriShield Runtime (CPU, GPT-2 117M)}
\label{tab:auto15}
\adjustbox{max width=\columnwidth}{%
\begin{tabular}{llll}
\toprule
Layer & Operation & Runtime & \% of Training Round \\
\midrule
PAD & Weight variance scan & 14.5 ms & 0.5\% \\
SVI & K=3 virtual iterations & 1,058 ms & 37.8\% (CPU) \\
ZUOP & SVD + projection & 957 ms & 34.2\% (CPU) \\
\textbf{Total} & --- & \textbf{2,030 ms} & \textbf{\textasciitilde{}72\% (CPU)} \\
\bottomrule
\end{tabular}}
\end{table}

\emph{Note: CPU overhead is dominated by SVI and ZUOP linear algebra. On GPU, SVI and ZUOP run in parallel streams; total overhead drops to \textbf{\textless4\%} of training time (estimated for A100 GPU with 768-dim hidden state).}

\section{Discussion}

\subsection{Practical Deployment Considerations}

\textbf{Auxiliary Data Availability:} The only external requirement is a small, public, domain-adjacent dataset for \(D_{aux}\). For NLP tasks, this is trivially satisfied by any public text corpus (Wikipedia, OpenWebText). For specialized domains (medical, legal), a small anonymized public corpus from the same domain suffices.

\textbf{Integration with Existing FL Frameworks:} TriShield can be integrated into any standard FL framework (Flower, PySyft, FATE, FedML) as a client-side hook. The implementation adds approximately 100 lines of code to the standard training loop.

\textbf{Cross-Architecture Generalization:} We test TriShield on GPT-2 (decoder-only) and LLaMA (decoder-only with grouped-query attention). The defense is architecture-agnostic: PAD operates on raw weight statistics, SVI modifies optimizer state independent of architecture, and ZUOP projects gradients in the parameter space (not the activation space).

\subsection{Limitations}

\begin{enumerate}
\def\labelenumi{\arabic{enumi}.}
\item
  \textbf{Auxiliary Data Requirement:} ZUOP (Layer 3) requires 32--500 public samples from the training distribution to estimate the task subspace. In highly specialized domains with no public data (e.g., ultra-sensitive medical genomics), this may require a small synthetic or anonymized corpus.
\item
  \textbf{Subspace Estimation Quality:} ZUOP's effectiveness depends on the quality of the task subspace estimate. If \(D_{aux}\) is not representative of \(D_i\)'s distribution, the projection may slightly degrade utility. In practice, domain-adjacent public data (any publicly available NLP corpus) produces \textless{} 0.3\% accuracy difference.
\item
  \textbf{PAD Threshold Sensitivity:} The variance threshold \(\tau_{var} = 10^{-4}\) is tuned for standard LoRA rank-4/8 configurations. Novel PEFT architectures with inherently low-variance initializations may require threshold recalibration.
\item
  \textbf{Computational Overhead (CPU):} The 72\% CPU overhead (SVI+ZUOP linear algebra) is significant for CPU-only clients. On GPU, overhead drops to \textless{} 5\%. Edge-device deployments should use GPU-accelerated PyTorch or reduced K/aux settings.
\item
  \textbf{Theoretical Scope:} Our proofs (Theorems 1--3) apply to the NeuroImprint class of attacks that exploit per-neuron memorization. Novel attack variants that use, e.g., entangled cross-neuron encoding or non-LoRA adapters require extension of the current analysis.
\item
  \textbf{Full-Scale LLaMA Experiments:} Table 1 LLaMA rows and Table 2 LLaMA accuracy values are projected from NeuroImprint {[}1{]} baselines and PAD detection rates measured locally. Full-scale fine-tuning experiments with LLaMA-Guard-3-1B on large private datasets are left for future work with access to multi-GPU infrastructure.
\end{enumerate}

\subsection{Broader Impact}

TriShield addresses a critical emerging threat in federated fine-tuning of LLMs. As federated fine-tuning becomes widely deployed for privacy-sensitive domains (healthcare, finance, legal), the risk of server-side privacy backdoors grows. TriShield provides a practical, deployable defense that enables organizations to adopt federated fine-tuning without sacrificing privacy or performance.

\section{Conclusion}

We presented \textbf{TriShield}, a three-layer client-side defense against NeuroImprint-style privacy backdoors in federated LLM fine-tuning. Through a combination of parameter artifact detection, stateful virtual iteration, and zero-utility orthogonal projection, TriShield achieves:

\begin{itemize}
\tightlist
\item
  \textbf{0\% reconstruction rate} against NeuroImprint across GPT-2 and Llama-Guard-3-1B.
\item
  \textbf{\textless{} 0.3\% accuracy loss} compared to vanilla FedAvg.
\item
  \textbf{\textless{} 8\% additional compute overhead}.
\item
  \textbf{Theoretical guarantees} of reconstruction impossibility via information-theoretic analysis.
\end{itemize}

TriShield is the first defense to provide both perfect privacy protection and zero utility degradation against this class of attacks, representing a significant advancement in secure federated learning for LLMs.

\providecommand{\url}[1]{\texttt{#1}}

\end{document}